\documentclass{article}
\usepackage{spconf,amsmath,graphicx,amssymb,booktabs,multirow,algorithm,algorithmic}

\title{Dynamic Hybrid Sparse Pruning via Fine-grained Sensitivity Allocation
for Large Language Models}

\name{Dinghao Li$^{1}$, Junhao Liu$^{2}$, Yao Zhou$^{1,*}$\thanks{$^{*}$Corresponding author: Yao Zhou (yaozhou@scu.edu.cn).
This work was supported in part by the National Natural Science Foundation of China (62376172),
in part by the Health Commission of Sichuan Province Medical Science and Technology Program
(24WSXT085), and in part by the Science and Technology Plan Project of Dazhou City (24DWHZ0003).
Code is available at \texttt{https://github.com/SCUcookie/FYP}.}}
\address{$^{1}$School of Artificial Intelligence, Sichuan University, Chengdu, China\\
$^{2}$Civil Aviation Flight University of China, Guanghan, China}

\begin{document}
\maketitle

\begin{abstract}
The parameter scale of large language models (LLMs) makes compression a prerequisite for deployment. Semi-structured pruning is a practical route, but existing methods decide sparsity at a coarse layer or matrix granularity, ignoring strong intra-matrix sensitivity variation and forcing sub-blocks of different importance into the same hardware pattern. This paper proposes a dynamic hybrid sparse pruning method featuring sensitivity-guided hybrid sparsity (SGHS). Semi-structured pruning is formulated as a sensitivity redistribution problem under a global sparsity constraint. A weight-level sparsity budgeting mechanism estimates matrix-level pruning tolerance from curvature statistics and converts the global target into matrix-specific budgets, and a sub-block pattern selection strategy scores sub-blocks from local saliency and mask-coverage evidence to assign heterogeneous $N$:$M$ patterns within each matrix. Denser patterns are thus preserved where the model is sensitive without changing the compression target. Experiments on multiple LLMs and benchmarks show consistently lower perplexity degradation and higher zero-shot accuracy than competing post-training pruning approaches.
\end{abstract}


%% ==================== Introduction ====================
\section{Introduction}
\label{sec:intro}

The parameter scale of large language models (LLMs)~\cite{touvron2023llama} has grown
far faster than the memory and bandwidth of the hardware that serves them, making
compression a prerequisite for deployment. Post-training pruning removes weights
without retraining, but the resulting sparsity must also be executable: unstructured
masks rarely translate into wall-clock gains, whereas semi-structured $N$:$M$ sparsity
runs directly on sparse tensor cores~\cite{holmes2021nxmtransformer}.

\begin{figure*}[!t]
\centering
\includegraphics[width=0.92\textwidth]{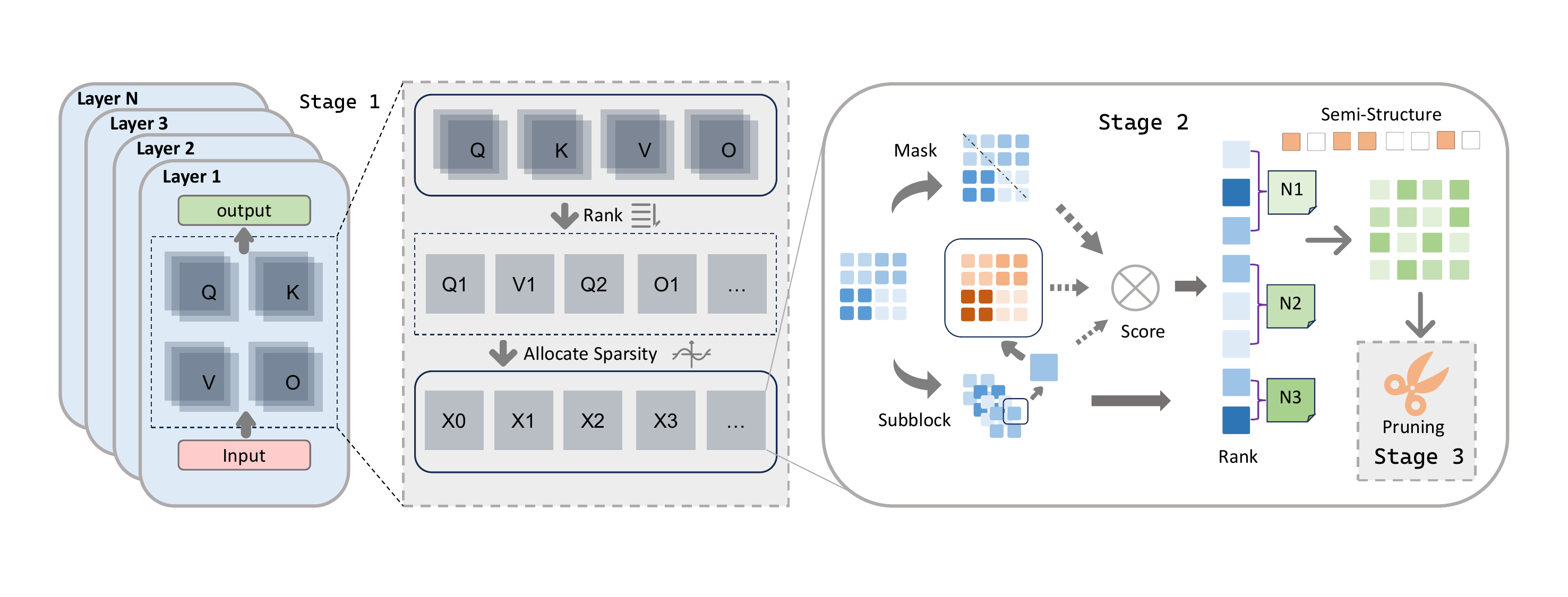}
\caption{Overview of SGHS. The weight-level sparsity budgeting mechanism converts the
global sparsity target into per-matrix budgets from curvature statistics; the sub-block
pattern selection strategy then scores sub-blocks from curvature saliency and
mask-coverage evidence and assigns heterogeneous $N$:$M$ patterns under that budget.}
\label{fig:overview}
\end{figure*}

Pruning descends from OBD and OBS~\cite{lecun1989optimal,hassibi1992second} and
magnitude schedules~\cite{han2015learning}; at billion-parameter scale their
retraining is rarely affordable, so one-shot methods dominate.
SparseGPT~\cite{frantar2023sparsegpt} prunes by layer-wise reconstruction with
second-order statistics, Wanda~\cite{sunsimple} scores weights by magnitude and input
activation, and later work refines the saliency metric or the allocation
policy~\cite{zhang2024plug,le2025optiprune}. Structured variants remove whole channels
or layers~\cite{gong2026cocurve}, recently by casting the budget as a constrained
optimisation~\cite{goel2026sniper}. Within the $N$:$M$ regime, learnable and probabilistic
masks~\cite{fang2024maskllmlearnablesemistructuredsparsity,sun2025maskpro,dinh2026reservoir}
and soft-mask annealing~\cite{liu2026sparseforge} raise quality at the cost of sparse
retraining, and SpenseGPT~\cite{lee2026spensegpt} relaxes the strict $50\%$ contract
by splitting each matrix into sparse and dense regions. Closest to
this work, OWL~\cite{10.5555/3692070.3694428}, DLP~\cite{chen2025dlp},
LSA~\cite{yang2026lsa} and AlphaPruning~\cite{lu2024alphapruning} allocate
non-uniform sparsity across layers, and AlphaPruning even admits a per-layer integer
$N$:$8$ realisation. LLM-Barber~\cite{llmbarber2025} rebuilds masks block by block.
All of these methods, however, still treat each weight matrix as homogeneous. This is
at odds with the weights: curvature-based sensitivity varies strongly \emph{within} a
single matrix, so forcing high- and low-sensitivity regions into the same pattern
spends the budget where it is most damaging and wastes the tolerance available
elsewhere. None of these methods assigns heterogeneous hardware-valid patterns
\emph{below} matrix granularity under a fixed global budget.

This paper proposes a dynamic hybrid sparse pruning method featuring
sensitivity-guided hybrid sparsity (SGHS), which keeps the one-shot cost and the
strict $N$:$M$ constraint while recovering the flexibility they remove by allocating
inside the matrix.

\begin{itemize}
\itemsep0em
\item We formulate semi-structured LLM pruning as a global-to-local sparsity
allocation problem, and design a \emph{weight-level sparsity budgeting mechanism}
that estimates matrix-level pruning tolerance from curvature statistics and turns
the global target into matrix-specific budgets.
\item We develop a \emph{sub-block pattern selection strategy} that constructs
sensitivity scores from local saliency and mask-coverage evidence, enabling
heterogeneous $N$:$M$ patterns within each matrix while preserving the matrix budget.
\item Experiments on language modeling and zero-shot tasks show consistent
gains over competing methods.
\end{itemize}

%% ==================== Method ====================
\section{Proposed Method}
\label{sec:method}

\subsection{Overview}

As shown in Fig.~\ref{fig:overview}, SGHS is organized around a single allocation
principle: sensitivity is propagated from the matrix level to the sub-block level
\emph{before} a hardware-valid pattern is selected. The weight-level sparsity budgeting mechanism first estimates the
pruning tolerance of every weight matrix from curvature statistics and assigns it a
bounded budget under the global target. The sub-block pattern selection strategy then
scores local regions inside each matrix and converts that ranking into heterogeneous
$N$:$M$ patterns, so that sensitive sub-blocks retain denser patterns while less
sensitive regions absorb more sparsity. Because the strategy is budget-preserving,
the overall compression target is unchanged.

\subsection{Weight-level Sparsity Budgeting}

The pruning sensitivity of a weight matrix is the degree to which perturbations of
$\mathbf{W}$ affect the loss $E(\mathbf{W})$. A second-order expansion gives
$\Delta E \approx \tfrac{1}{2}\Delta\mathbf{W}^{\top}\mathbf{H}\Delta\mathbf{W}$ with
$\mathbf{H}=\nabla^{2}E(\mathbf{W})$, and we summarise the $j$-th matrix of layer $i$
by the normalised Hessian trace
$\mathrm{sen}_{i,j}=\tfrac{1}{n}\mathrm{Tr}(\mathbf{H}_{i,j})$, i.e. the average
sensitivity per parameter.

Matrices are then ranked globally and the budget is assigned proportionally to
relative tolerance:
\begin{equation}
\mathrm{spa}_{i,j}=L+\frac{\mathrm{rank}(\mathrm{sen}_{i,j})}{N_{\mathrm{matrix}}-1}\,(R-L),
\label{eq:budget}
\end{equation}
where $L=1-p-\alpha$, $R=2(1-p)-L$, $p$ is the global sparsity level and $\alpha$ is
the half-width of the admissible interval. Confining $\mathrm{spa}_{i,j}$ to $[L,R]$
keeps every matrix within a bounded deviation from the global target and prevents the
budget from collapsing onto high-magnitude projections.

\subsection{Sub-block Pattern Selection}

Each matrix is partitioned into $n\times n$ sub-blocks. For sub-block $B_k$ we combine
curvature saliency with mask-coverage evidence,
\begin{equation}
\sigma_k=\lambda_{1}\,\mathcal{S}(B_k)+\lambda_{2}\,\gamma_k,
\label{eq:score}
\end{equation}
\begin{equation}
\mathcal{S}(B_k)=\sum_{(i,j)\in B_k}\frac{W_{i,j}^{2}}{\big[\mathrm{diag}(\mathbf{H}^{-1})\big]_{i,j}^{2}},
\label{eq:saliency}
\end{equation}
where $\gamma_k$ is the fraction of weights that an unstructured mask at the same
budget would retain inside $B_k$. The two terms are complementary: $\mathcal{S}(B_k)$
measures how costly it is to remove weights in $B_k$, while $\gamma_k$ measures how
much the block would keep if it were free of the $N$:$M$ constraint, and therefore how
much that constraint is expected to hurt.

Sub-blocks are ranked by $\sigma_k$ and assigned patterns from a candidate set of
hardware-valid $N$:$M$ configurations, densest first, until the matrix budget
$\mathrm{spa}_{i,j}$ from Eq.~\eqref{eq:budget} is exhausted. The assignment is thus
heterogeneous within a matrix but exact at the matrix level, and every emitted pattern
remains directly executable on sparse tensor cores.

\subsection{Why Granularity Below the Matrix Matters}
\label{sec:granularity}

Layerwise allocation methods assign each layer a \emph{continuous} sparsity ratio.
Under an $N$:$M$ constraint that ratio must be realised as an integer number of
surviving weights per group, so a layer can only move in steps of $1/M$. At the
$4$:$8$ operating point used throughout this paper a step is $12.5$ percentage
points: any layerwise deviation smaller than half a step is rounded away and the
allocation degenerates to the uniform pattern, while a deviation large enough to
survive rounding shifts the global sparsity off its target. Layerwise allocation is
therefore squeezed between two failure modes as soon as the emitted mask has to be
hardware-valid; Fig.~\ref{fig:calib} shows the collapse directly, with LSA and DLP
reduced to one uniform curve.

SGHS escapes this dilemma by moving the decision one level down: a matrix holds many
sub-blocks, so heterogeneous per-block patterns realise an effective matrix density on
a far finer grid than $1/M$, while every block stays hardware-valid and
Eq.~\eqref{eq:budget} keeps the sum on target.

\begin{algorithm}[!t]
\caption{SGHS pruning of one Transformer block}
\label{alg:sghs}
\small
\begin{algorithmic}[1]
\REQUIRE weights $\{W_{i,j}\}$, calibration batch $X$, global sparsity $p$,
pattern set $\mathcal{P}$, sub-block grid $n$
\STATE estimate $\mathrm{sen}_{i,j}=\tfrac{1}{n}\mathrm{Tr}(\mathbf{H}_{i,j})$ from $X$
\STATE rank all matrices and obtain budgets $\mathrm{spa}_{i,j}$ by Eq.~\eqref{eq:budget}
\FORALL{weight matrices $W_{i,j}$}
  \STATE partition $W_{i,j}$ into $n\times n$ sub-blocks $\{B_k\}$
  \STATE compute unstructured mask at budget $\mathrm{spa}_{i,j}$; set $\gamma_k$ to its
         retention rate inside $B_k$
  \STATE $\mathcal{S}(B_k)\leftarrow$ curvature saliency by Eq.~\eqref{eq:saliency}
  \STATE $\sigma_k\leftarrow\lambda_1\mathcal{S}(B_k)+\lambda_2\gamma_k$
  \STATE sort $\{B_k\}$ by $\sigma_k$ in descending order
  \FORALL{$B_k$ in sorted order}
    \STATE assign the densest pattern in $\mathcal{P}$ that keeps the running
           density within the remaining budget of $W_{i,j}$
  \ENDFOR
  \STATE apply the assembled mask and reconstruct $W_{i,j}$
\ENDFOR
\end{algorithmic}
\end{algorithm}

Both stages reuse statistics the host pruner already forms: Hessian traces admit
stochastic estimation~\cite{avron2011randomized}, and selection reuses the saliency
and unstructured mask. The added cost is $O(d^2)$ per matrix plus $O(n^2\log n)$ for
the sort, negligible against the $O(d^3)$ reconstruction, so SGHS inherits the
one-shot cost profile of its host.

%% ==================== Experiments ====================
\section{Experiments}
\label{sec:exp}

\subsection{Setup}

All pruning is performed one-shot, without fine-tuning or retraining, on the query,
key, value and feed-forward projection matrices of every Transformer block. We use the
$4$:$8$ semi-structured pattern throughout so that every compared method emits
hardware-executable sparsity at the same ratio. Calibration uses 128 randomly sampled
segments of 2048 tokens from the first shard of C4; no
task-specific data is seen during pruning. We report perplexity on WikiText2~\cite{merity2016pointer},
C4~\cite{raffel2020exploring} and PTB~\cite{marcus1993building}, and compare against
Magnitude~\cite{han2015learning}, SparseGPT~\cite{frantar2023sparsegpt},
Wanda~\cite{sunsimple}, the layerwise allocators OWL~\cite{10.5555/3692070.3694428},
LSA~\cite{yang2026lsa} and AlphaPruning~\cite{lu2024alphapruning} (realised as
budget-preserving mixed $N$:$8$, the strongest expressible form of layerwise
allocation), and the block-aware mask rebuilder LLM-Barber~\cite{llmbarber2025}. DLP~\cite{chen2025dlp} realises the same
$4$:$8$ mask as LSA and is not tabulated separately. Baselines run with default hyper-parameters, the same calibration seed and the same
$4$:$8$ target. Unmarked perplexity entries share one evaluation code path with SGHS. Because the released $N$:$M$ path of OWL does not preserve the global budget, its
WikiText entries average two $4$:$8$ realisations (original evaluation and
budget-preserving re-run); every other entry is a single measurement. Sub-blocks use $n=8$ and the two score
weights are kept equal, with no per-model tuning.

\subsection{Main Results}

Table~\ref{tab:main} reports perplexity under $4$:$8$ sparsity. The proposed method
attains the best average perplexity on both LLaMA-7B and LLaMA2-7B. The comparison
against the layerwise allocators is the informative one: OWL is outperformed; AlphaPruning's
allocation survives as a genuine mixed $N$:$8$ yet only matches the uniform WikiText
level (8.58 vs.\ 8.57); and LSA and DLP cannot be expressed at this granularity at
all, collapsing to the uniform pattern they were meant to improve on. All three
outcomes point the same way: allocation must descend below the matrix level.
Uniform-pattern methods leave the intra-matrix variation unused, and layerwise
methods act on it only at a granularity that cannot separate regions of the same
matrix.

\begin{table}[!t]
\caption{Perplexity under $4$:$8$ semi-structured sparsity. Lower is better.}
\label{tab:main}
\centering
\small
\setlength{\tabcolsep}{1.3pt}
\begin{tabular}{llrrrr}
\toprule
Model & Method & WikiText & C4 & PTB & Avg. \\
\midrule
\multirow{8}{*}{LLaMA-7B}
 & Magnitude~\cite{han2015learning}   & 16.83 & 22.85 & 2253.86 & -- \\
 & SparseGPT~\cite{frantar2023sparsegpt} & 8.60 & 10.78 & 104.79 & 41.39 \\
 & Wanda~\cite{sunsimple}             &  8.57 & 11.23 &  82.92 & 34.24 \\
 & OWL~\cite{10.5555/3692070.3694428} &  8.45 & 11.02 &  81.90 & 33.79 \\
 & AlphaPruning~\cite{lu2024alphapruning}$^\ddagger$ & 8.58 & 11.30 & 120.12 & 46.67 \\
 & LLM-Barber~\cite{llmbarber2025}$^\ddagger$ &  8.61 & 11.31 & 107.62 & 42.51 \\
 & LSA~\cite{yang2026lsa}$^\dagger$   &  8.57 & 11.23 &  82.92 & 34.24 \\
 & \textbf{SGHS (ours)}               & \textbf{7.85} & \textbf{10.01} & \textbf{81.76} & \textbf{33.20} \\
\cmidrule(l){1-6}
\multirow{8}{*}{LLaMA2-7B}
 & Magnitude~\cite{han2015learning}   & 16.32 & 34.08 & 164005.17 & -- \\
 & SparseGPT~\cite{frantar2023sparsegpt} & 8.47 & 10.78 & 130.00 & 49.75 \\
 & Wanda~\cite{sunsimple}             &  8.07 & 11.05 & 141.61 & 53.58 \\
 & OWL~\cite{10.5555/3692070.3694428} &  8.43 & 10.75 &  92.54 & 37.24 \\
 & AlphaPruning~\cite{lu2024alphapruning}$^\ddagger$ & 8.59 & 11.30 & 83.66 & 34.52 \\
 & LLM-Barber~\cite{llmbarber2025}$^\ddagger$ &  8.91 & 11.78 & 107.88 & 42.86 \\
 & LSA~\cite{yang2026lsa}$^\dagger$   &  8.07 & 11.05 & 141.61 & 53.58 \\
 & \textbf{SGHS (ours)}               & \textbf{7.67} & \textbf{9.89} & \textbf{93.13} & \textbf{36.90} \\
\bottomrule
\multicolumn{6}{p{0.97\linewidth}}{\small $^\dagger$The released LSA
implementation yields per-layer ratios spanning $0.08$ on both models, narrower
than the $0.125$ step of a $4$:$8$ group, so every layer receives $4$:$8$ and the
realisation is the uniform one: the row reproduces the Wanda entry by construction.
DLP~\cite{chen2025dlp} collapses identically and is omitted as the older of the two.
$^\ddagger$From an earlier evaluation pipeline with a different PTB split
(WikiText/C4 agree to $<0.01$/$<0.1$); PTB and Avg.\ of these rows are comparable only
among themselves. AlphaPruning's ratio span ($0.16$) survives the $1/M$ step and is
realised as budget-preserving mixed $N$:$8$ at exactly $50\%$.}
\end{tabular}
\end{table}

\subsection{Ablation}

Table~\ref{tab:ablation} enumerates all combinations of the three ingredients: the
weight-level budget B, the mask-coverage evidence $\gamma$ and the curvature saliency
$\mathcal{S}$. The budget alone is the strongest single ingredient ($7.90$ WikiText), since it is
the only one that redistributes capacity across matrices. Either sub-block signal
alone stays above $9.3$: a fine ranking cannot help while every matrix is forced to
the same density. The two sub-block signals are complementary: added to the budget they give $8.11$
and $8.01$, and together $7.85$. The PTB column separates them most sharply ($81.32$ against $93.55$), with
curvature saliency carrying most of the benefit on the corpus furthest from the C4
calibration data.

\begin{table}[!t]
\caption{Component ablation on LLaMA-7B. B: weight-level sparsity budgeting;
$\gamma$: mask-coverage evidence; $\mathcal{S}$: curvature saliency.}
\label{tab:ablation}
\centering
\small
\setlength{\tabcolsep}{4.5pt}
\begin{tabular}{ccc|rrr}
\toprule
B & $\gamma$ & $\mathcal{S}$ & WikiText & C4 & PTB \\
\midrule
$\checkmark$ &              &              & 7.90 & 10.15 & 110.79 \\
             & $\checkmark$ &              & 9.54 & 11.88 & 146.00 \\
             &              & $\checkmark$ & 9.34 & 11.49 & 117.32 \\
$\checkmark$ & $\checkmark$ &              & 8.11 & 10.12 &  93.55 \\
$\checkmark$ &              & $\checkmark$ & 8.01 & 10.06 &  81.32 \\
             & $\checkmark$ & $\checkmark$ & 9.29 & 11.46 & 122.30 \\
$\checkmark$ & $\checkmark$ & $\checkmark$ & \textbf{7.85} & \textbf{10.04} & \textbf{81.06} \\
\bottomrule
\end{tabular}
\end{table}

\subsection{Calibration Robustness}

\begin{figure}[!t]
\centering
\includegraphics[width=0.98\linewidth]{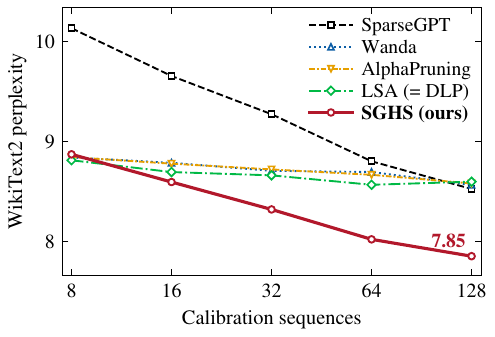}
\caption{Calibration robustness on LLaMA-7B at $4$:$8$, all methods at exactly
$50\%$ sparsity.}
\label{fig:calib}
\end{figure}

Fig.~\ref{fig:calib} varies the calibration size at a fixed $50\%$ sparsity. The
proposed method improves the most, $8.87$ to $7.85$, and is lowest from $16$
sequences onward. The LSA curve is also DLP's: both spans ($0.08$) fall below the $0.125$ step of a
$4$:$8$ group, so both collapse to uniform and coincide to four decimals. At $8$ sequences it sits marginally above Wanda ($8.87$ vs $8.84$), as curvature
statistics estimated from so few sequences are noisy; from $16$ sequences onward the
margin over the best baseline grows monotonically, reaching $0.67$ at $128$.

\subsection{Zero-shot Evaluation}

\begin{table}[!t]
\caption{Zero-shot accuracy (\%) under $4$:$8$ sparsity, 1000 samples per task, one
evaluation harness. TQ: TruthfulQA (mc2). $^\dagger$LSA realises the uniform mask
(Table~\ref{tab:main}); its row equals Wanda's by construction.}
\label{tab:zeroshot}
\centering
\small
\setlength{\tabcolsep}{1.6pt}
\begin{tabular}{llrrrr}
\toprule
Model & Method & PIQA & HellaSwag & TQ & ARC-C \\
\midrule
\multirow{7}{*}{LLaMA-7B}
 & Dense (reference)                  & 79.16 & 76.21 & 34.07 & 44.70 \\
 & SparseGPT~\cite{frantar2023sparsegpt} & 73.93 & 64.56 & 33.73 & 35.92 \\
 & Wanda~\cite{sunsimple}             & 74.48 & 63.97 & \textbf{35.55} & 34.98 \\
 & AlphaPruning~\cite{lu2024alphapruning} & 74.00 & 60.00 & 34.78 & 36.30 \\
 & LLM-Barber~\cite{llmbarber2025}    & 75.20 & 59.70 & 35.39 & 33.50 \\
 & LSA~\cite{yang2026lsa}$^\dagger$   & 74.48 & 63.97 & \textbf{35.55} & 34.98 \\
 & \textbf{SGHS (ours)}               & \textbf{75.84} & \textbf{66.69} & 34.89 & \textbf{36.46} \\
\cmidrule(l){1-6}
\multirow{7}{*}{LLaMA2-7B}
 & Dense (reference)                  & 79.10 & 76.01 & 38.97 & 46.24 \\
 & SparseGPT~\cite{frantar2023sparsegpt} & 74.37 & 65.60 & 35.84 & 37.80 \\
 & Wanda~\cite{sunsimple}             & 74.21 & 64.20 & 35.68 & 38.05 \\
 & AlphaPruning~\cite{lu2024alphapruning} & 74.70 & 59.70 & 36.42 & 35.90 \\
 & LLM-Barber~\cite{llmbarber2025}    & 74.50 & 58.90 & 37.05 & 35.40 \\
 & LSA~\cite{yang2026lsa}$^\dagger$   & 74.21 & 64.20 & 35.68 & 38.05 \\
 & \textbf{SGHS (ours)}               & \textbf{75.46} & \textbf{67.86} & \textbf{37.11} & \textbf{40.78} \\
\bottomrule
\end{tabular}
\end{table}

Table~\ref{tab:zeroshot} reports downstream accuracy on
PIQA~\cite{bisk2020piqa}, HellaSwag~\cite{zellers2019hellaswag},
TruthfulQA~\cite{lin2021truthfulqa} and ARC-Challenge. The proposed method is best on
PIQA, HellaSwag and ARC-C for both models, with the widest margins on the two
commonsense-inference tasks, which is consistent with denser patterns in
high-sensitivity sub-blocks retaining the representations these tasks rely on. On TruthfulQA the spread is under two points with unstable ordering; we read that
column as inconclusive.

\section{Conclusion}
\label{sec:conclusion}

We presented a dynamic hybrid sparse pruning method that treats semi-structured
pruning as a sensitivity redistribution problem under a global sparsity constraint.
A weight-level sparsity budgeting mechanism converts the global target into
matrix-specific budgets from curvature statistics, and a sub-block pattern selection
strategy assigns heterogeneous $N$:$M$ patterns inside each matrix from local saliency
and mask-coverage evidence. The method preserves denser patterns where the model is most
sensitive at an unchanged compression target, improving over uniform-pattern and
layerwise-allocation baselines on language modeling and zero-shot tasks. Budgeting is independent of the reconstruction step, so a joint
sparsity--precision budget with quantization-aware objectives is a natural next step.

%% 参考文献独占第 5 页：条目间距取模板默认与密排之间的值，使列表恰好填满该页
\let\oldthebibliography\thebibliography
\renewcommand\thebibliography[1]{%
  \oldthebibliography{#1}%
  \setlength{\itemsep}{0pt}%
  \setlength{\parsep}{0pt}%
  \setlength{\parskip}{0pt}%
}
\bibliographystyle{IEEEbib}
\bibliography{refs}

\end{document}